\documentclass{article}
\usepackage[T1]{fontenc}
\usepackage{amssymb}
\usepackage{colacl}
\usepackage{epsfig}
\usepackage{verbatim}

\newcommand{\argmax}{\mathop{\rm argmax}}

\title{Exploiting Diversity in Natural Language Processing:\\
  Combining Parsers}
\author{John C. Henderson and Eric Brill\\
  Department of Computer Science\\
  Johns Hopkins University\\
  Baltimore, MD 21218\\
  \{jhndrsn,brill\}@cs.jhu.edu
}

\begin{document}

\maketitle

\begin{abstract}
 \begin{picture}(0,0)
 \put(0,215){Appears in}
 \put(5,200){Proceedings of the Fourth Conference on Empirical Methods
 in Natural Language Processing (EMNLP-99),}
 \put(10,185){pages 187--194.  College Park, Maryland,
 USA.  June, 1999.} 
 \end{picture}
Three state-of-the-art statistical parsers are combined to produce
more accurate parses, as well as new bounds on achievable Treebank
parsing accuracy.  Two general approaches are presented and two
combination techniques are described for each approach.  Both
parametric and non-parametric models are explored.
%
The resulting parsers surpass the best previously published
performance results for the Penn Treebank. 
\vspace{0.38in}
\end{abstract}

\section{Introduction}

The natural language processing community is in the strong position of
having many available approaches to solving some of its most
fundamental problems.
The machine learning community has been in a similar situation and has
studied the combination of multiple classifiers
\cite{wolpert:stacking,heath96:committee}.  Their theoretical finding
is simply stated: classification error rate decreases toward the noise
rate exponentially in the number of independent, accurate classifiers.
The theory has also been validated empirically.

Recently, combination techniques have been investigated for part of
speech tagging with positive results
\cite{halteren98:combine,brillwu:combine}.  In both cases the
investigators were able to achieve significant improvements over the
previous best tagging results.  Similar advances have been made in
machine translation \cite{mt:threeheads}, speech recognition
\cite{rover} and named entity recognition \cite{borthwick98:named}.

The corpus-based statistical parsing community has many fast and
accurate automated parsing systems, including systems produced by
\newcite{collins:parsing97}, \newcite{charniak:parsing} and
\newcite{adwait:parsing}.
These three parsers have given the best reported parsing results on
the Penn Treebank Wall Street Journal corpus \cite{marcus:treebank}.
We used these three parsers to explore parser combination
techniques.

\section{Techniques for Combining Parsers}
\subsection{Parse Hybridization}
\label{section:selection}

We are interested in combining the substructures of the input parses
to produce a better parse.  We call this approach {\em parse
  hybridization}.  The substructures that are unanimously hypothesized
by the parsers should be preserved after combination, and the
combination technique should not foolishly create substructures for
which there is no supporting evidence.  These two principles guide
experimentation in this framework, and together with the evaluation
measures help us decide which specific type of substructure to
combine.

The precision and recall measures (described in more detail in Section
\ref{section:precisionrecall}) used in evaluating Treebank parsing
treat each constituent as a separate entity, a minimal unit of
correctness.  Since our goal is to perform well under these measures
we will similarly treat constituents as the minimal substructures for
combination.

\subsubsection{Constituent Voting}

One hybridization strategy is to let the parsers vote on constituents'
membership in the hypothesized set.  If enough parsers suggest that a
particular constituent belongs in the parse, we include it.  We call
this technique {\em constituent voting}.  We include a constituent in
our hypothesized parse if it appears in the output of a majority of
the parsers.  In our particular case the majority requires the
agreement of only two parsers because we have only three.  This
technique has the advantage of requiring no training, but it has the
disadvantage of treating all parsers equally even though they may have
differing accuracies or may specialize in modeling different
phenomena. 

\subsubsection{Na\"ive Bayes}

Another technique for parse hybridization is to use a na\"ive Bayes
classifier to determine which constituents to include in the parse.
The development of a na\"ive Bayes
classifier involves learning how much each parser should be trusted
for the decisions it makes.
Our original hope in combining these parsers is that their errors are
independently distributed.
This is equivalent to the assumption used in probability
estimation for na\"ive Bayes classifiers, namely that the attribute
values are conditionally independent when the target value is given.
For this reason, na\"ive Bayes classifiers are well-matched to this problem.

In Equations \ref{eqn:bayesformulation} through
\ref{eqn:bayesmaximization} we develop the model for constructing our
parse using na\"ive Bayes classification.
 $\mathcal C$ is the union of the
sets of constituents suggested by the parsers.
$\pi(c)$ is a binary function
returning $t$ (for $true$) precisely when the constituent
$c \in \mathcal C$
should be included in the hypothesis.
$M_i(c)$ is a binary function returning $t$ when parser $i$
(from among the $k$ parsers) suggests constituent $c$ should be in
the parse.
The hypothesized parse is then the set of constituents
that are likely ($P>0.5$) to be in the parse according to this model.
\begin{eqnarray}
\lefteqn{\argmax\limits_{\pi(c)}  P(\pi(c) | M_1(c)\ldots M_k(c))}\nonumber\\
&=&\argmax\limits_{\pi(c)}\frac{P(M_1(c)\ldots M_k(c)|\pi(c))P(\pi(c))}{P(M_1(c)\ldots M_k(c))}
  \label{eqn:bayesformulation}
  \\
  & = &
  \argmax\limits_{\pi(c)}  P(\pi(c))\prod_{i=1}^k{\frac{P(M_i(c)|\pi(c))}{P(M_i(c))}}
  \label{eqn:bayesindependence}
  \\
  & = &
  \argmax\limits_{\pi(c)}  P(\pi(c))\prod_{i=1}^k{P(M_i(c)|\pi(c))}
  \label{eqn:bayesmaximization}
\end{eqnarray}
The estimation of the probabilities in the model is carried out as
shown in Equation \ref{eqn:bayescounts}.  Here $N(\cdot)$ counts the
number of hypothesized constituents in the development set that match
the binary predicate specified as an argument.
\begin{eqnarray}
  \lefteqn{P(\pi(c)=t)\prod_{i=1}^k{P(M_i(c)|\pi(c)=t)}} \nonumber\\
    & = &
    \frac{N(\pi(c)=t)}
    {|\mathcal C|}
    \prod_{i=1}^k{\frac{N(M_i(c),\pi(c)=t)}{N(\pi(c)=t)}}
    \label{eqn:bayescounts}
\end{eqnarray}

\subsubsection{Lemma: No Crossing Brackets}
\label{section:treestructure}

Under certain conditions the constituent voting and na\"ive Bayes
constituent combination techniques are guaranteed to produce sets of
constituents with no crossing brackets.  There are simply not enough
votes remaining to allow any of the crossing structures to enter the
hypothesized constituent set.

{\bf Lemma:} If the number of votes required by constituent voting is
greater than half of the parsers under consideration the resulting
structure has no crossing constituents.

{\bf Proof:} Assume a pair of crossing constituents appears in the
output of the constituent voting technique using $k$ parsers.  Call
the crossing constituents $A$ and $B$.  $A$ receives $a$ votes, and
$B$ receives $b$ votes.  Each of the constituents must have received
at least $\lceil\frac{k+1}{2}\rceil$ votes from the $k$ parsers, so $a
\geq \lceil\frac{k+1}{2}\rceil$ and $b \geq
\lceil\frac{k+1}{2}\rceil$.  Let $s = a+b$.  None of the parsers
produce parses with crossing brackets, so none of them votes for both
of the assumed constituents.  Hence, $s \leq k$.  But by addition of
the votes on the two parses, $s \geq 2\lceil\frac{k+1}{2}\rceil > k$,
a contradiction.\hfill $\blacksquare$ 

Similarly, when the na\"ive Bayes classifier is configured such that
the constituents require estimated probabilities strictly larger than
0.5 to be accepted, there is not enough probability mass remaining on
crossing brackets for them to be included in the hypothesis.

\subsection{Parser Switching}

In general, the lemma of the previous section does not ensure that all
the productions in the combined parse are found in the grammars of the
member parsers.  There is a guarantee of no crossing brackets but
there is no guarantee that a constituent in the tree has the same
children as it had in any of the three original parses.  One can
trivially create situations in which strictly binary-branching trees
are combined to create a tree with only the root node and the terminal
nodes, a completely flat structure.

This drastic tree manipulation is not appropriate for situations in
which we want to assign particular structures to sentences.  For
example, we may have semantic information (e.g. database query
operations) associated with the productions in a grammar.  If the
parse contains productions from outside our grammar the machine has no
direct method for handling them (e.g. the resulting database query may
be syntactically malformed).

We have developed a general approach for combining parsers when
preserving the entire structure of a parse tree is important.  The
combining algorithm is presented with the candidate parses and asked
to choose which one is best.  The combining technique must act as a
multi-position switch indicating which parser should be trusted for
the particular sentence.  We call this approach {\em parser
  switching}.  Once again we present both a non-parametric and a
parametric technique for this task.

\subsubsection{Similarity Switching}

First we present the non-parametric version of parser switching,
{\em similarity switching}:

\begin{enumerate}
\item
  From each candidate parse, $\pi_i$, for a sentence, create the
  constituent set $S_i$ by converting each constituent into its tuple
  representation.
\item
  The score for $\pi_i$ is $\sum\limits_{j\neq i} |S_j \cap S_i|$, where $j$
  ranges over the candidate parses for the sentence.
\item
  Switch to (use) the parser with the highest score for the sentence.
  Ties are broken arbitrarily. 
\end{enumerate}

The intuition for this technique is that we can measure a similarity
between parses by counting the constituents they have in common.  We
pick the parse that is most similar to the other parses by
choosing the one with the highest sum of pairwise similarities.  This
is the parse that is closest to the centroid of the observed parses
under the similarity metric.

\subsubsection{Na\"ive Bayes}
 
The probabilistic version of this procedure is straightforward.  We
once again assume independence among our various member parsers.
Furthermore, we know one of the original parses will be the
hypothesized parse, so the direct method of determining which one is
best is to compute the probability of each of the candidate parses
using the probabilistic model we developed in Section
\ref{section:selection}.  We model each parse as the decisions made to
create it, and model those decisions as independent events.  Each
decision determines the inclusion or exclusion of a candidate
constituent.  The set of candidate constituents comes from the union
of all the constituents suggested by the member parsers.  This is
summarized in Equation \ref{eqn:bayesswitch}.  The computation of
$P(\pi_i(c)|M_1\ldots M_k(c))$ has been sketched before in Equations
\ref{eqn:bayesformulation} through \ref{eqn:bayescounts}.  In this
case we are interested in finding the maximum probability parse,
$\pi_i$, and $M_i$ is the set of relevant (binary) parsing decisions
made by parser $i$. $\pi_i$ is a parse selected from among the outputs
of the individual parsers.  It is chosen such that the decisions it
made in including or excluding constituents are most probable under
the models for {\em all} of the parsers.
\begin{eqnarray}
  \lefteqn{\argmax\limits_{\pi_i} P(\pi_i|M_1\ldots M_k)} \nonumber\\
  & = &
  \argmax\limits_{\pi_i} \prod_{c}{P(\pi_i(c)|M_1(c)\ldots M_k(c))}
  \label{eqn:bayesswitch}
\end{eqnarray}

\section{Experiments}


The three parsers were trained and tuned by their creators on various
sections of the WSJ portion of the Penn Treebank, leaving only
sections 22 and 23 completely untouched during the development of any
of the parsers.  We used section 23 as the development set for our
combining techniques, and section 22 only for final testing. The
development set contained 44088 constituents in 2416 sentences and the
test set contained 30691 constituents in 1699 sentences. A sentence
was withheld from section 22 because its extreme length was
troublesome for a couple of the parsers.\footnote{The sentence in
  question was more than 100 words in length and included nested
  quotes and parenthetical expressions.
}

\label{section:precisionrecall}
The standard measures for evaluating Penn Treebank parsing performance
are precision and recall of the predicted constituents.  Each parse is
converted into a set of constituents represented as a tuples: (label,
start, end).  The set is then compared with the set generated from the
Penn Treebank parse to determine the precision and recall.  {\bf
  Precision} is the portion of hypothesized constituents that are
correct and {\bf   recall} is the portion of the Treebank constituents
that are hypothesized.


For our experiments
we also report the mean of precision and recall, which we denote by
$(P+R)/2$ and F-measure.  F-measure is the harmonic mean of precision
and recall, $2PR/(P+R)$.  It is closer to the smaller value of
precision and recall when there is a large skew in their values.

We performed three experiments to evaluate our techniques.  The first
shows how constituent features and context do not help in deciding
which parser to trust.  We then show that the combining techniques
presented above give better parsing accuracy than any of the
individual parsers.  Finally we show the combining techniques degrade
very little when a poor parser is added to the set.

\subsection{Context}

It is possible one could produce better models by introducing features
describing constituents and their contexts because one parser could be
much better than the majority of the others in particular situations.
For example, one parser could be more accurate at predicting noun
phrases than the other parsers.  None of the models we have presented
utilize features associated with a particular constituent (i.e. the
label, span, parent label, etc.) to influence parser preference.  This
is not an oversight.  Features and context were initially introduced
into the models, but they refused to offer any gains in performance.
While we cannot prove there are no such useful features on which one
should condition trust, we can give some insight into why the features
we explored offered no gain.

Because we are working with only three parsers, the only situation in
which context will help us is when it can indicate we should choose to
believe a single parser that disagrees with the majority hypothesis
instead of the majority hypothesis itself.  This is the only important
case, because otherwise the simple majority combining technique would
pick the correct constituent.  One side of the decision making process
is when we choose to believe a constituent should be in the parse,
even though only one parser suggests it.  We call such a constituent
an {\em isolated constituent}.  If we were working with more than
three parsers we could investigate {\em minority constituents}, those
constituents that are suggested by at least one parser, but which the
majority of the parsers do not suggest.

Adding the isolated constituents to our hypothesis parse could
increase our expected recall, but in the cases we investigated it
would invariably hurt our precision more than we would gain on recall.
Consider for a set of constituents the {\em isolated constituent
  precision} parser metric, the portion of isolated constituents that
are correctly hypothesized.  When this metric is less than 0.5, we
expect to incur more errors\footnote{This is in absolute terms, total
  errors being the sum of precision errors and recall errors.} than we
will remove by adding those constituents to the parse.

\begin{table*}
  \begin{center}
    \begin{tabular}{|l|rr|rr|rr|}
      \hline 
      \multicolumn{1}{|c|}{Constituent}
      & \multicolumn{2}{c|}{Parser1}
      & \multicolumn{2}{c|}{Parser2}
      & \multicolumn{2}{c|}{Parser3} 
      \\ 
      \multicolumn{1}{|c|}{Label}
      & \multicolumn{1}{c}{count} & \multicolumn{1}{c|}{Precision}
      & \multicolumn{1}{c}{count} & \multicolumn{1}{c|}{Precision}
      & \multicolumn{1}{c}{count} & \multicolumn{1}{c|}{Precision}
      \\
      \hline 
      ADJP    & 132   & 28.78 & 215   & 21.86  & 173   & 34.10 \\
      ADVP    & 150   & 25.33 & 129   & 21.70  & 102   & 31.37 \\
      CONJP   & 2     & 50.00 & 8     & 37.50  & 3     & 0.00 \\
      FRAG    & 51    & 3.92  & 29    & 27.58  & 11    & 9.09 \\
      INTJ    & 3     & 66.66 & 1     &100.00  & 2     & 50.00 \\
      LST     & 0     & NA    & 0     & NA     & 0     & NA \\
      NAC     & 0     & NA    & 13    & 53.84  & 7     & 14.28 \\
      NP      & 1489  & 21.08 & 1550  & 18.38  & 1178  & 27.33 \\
      NX      & 7     & 85.71 & 9     & 22.22  & 3     & 0.00 \\
      PP      & 732   & 23.63 & 643   & 20.06  & 503   & 27.83 \\
      PRN     & 20    & 55.00 & 33    & 54.54  & 38    & 15.78 \\
      PRT     & 12    & 16.66 & 20    & 40.00  & 16    & 37.50 \\
      QP      & 21    & 38.09 & 34    & 44.11  & 76    & 14.47 \\
      RRC     & 1     & 0.00  & 1     & 0.00   & 2     & 0.00 \\
      S       & 757   & 13.73 & 482   & 23.65  & 434   & 38.94 \\
      SBAR    & 331   & 11.78 & 196   & 23.97  & 178   & 34.83 \\
      SBARQ   & 0     & NA    & 6     & 16.66  & 3     & 0.00 \\
      SINV    & 3     & 66.66 & 11    & 81.81  & 13    & 30.76 \\
      SQ      & 2     & 0     & 11    & 18.18  & 3     & 33.33 \\
      UCP     & 6     & 16.66 & 12    & 8.33   & 8     & 12.50 \\
      VP      & 868   & 13.36 & 630   & 24.12  & 477   & 35.42 \\
      WHADJP  & 0     & NA    & 0     & NA     & 1     & 0.00 \\
      WHADVP  & 2     &100.00 & 5     & 40.00  & 1     & 100.00 \\
      WHNP    & 33    & 33.33 & 8     & 25.00  & 17    & 58.82 \\
      WHPP    & 0     & NA    & 0     & NA     & 2     & 100.00 \\
      X       & 0     & NA    & 2     & 100.00 & 1     & 0.00 \\
      \hline 
    \end{tabular}
    \caption{Isolated Constituent Precision By Constituent Label}
    \label{table:isolated:tag}
  \end{center}
\end{table*}
\begin{figure*}
  \begin{center}
    \epsfig{file=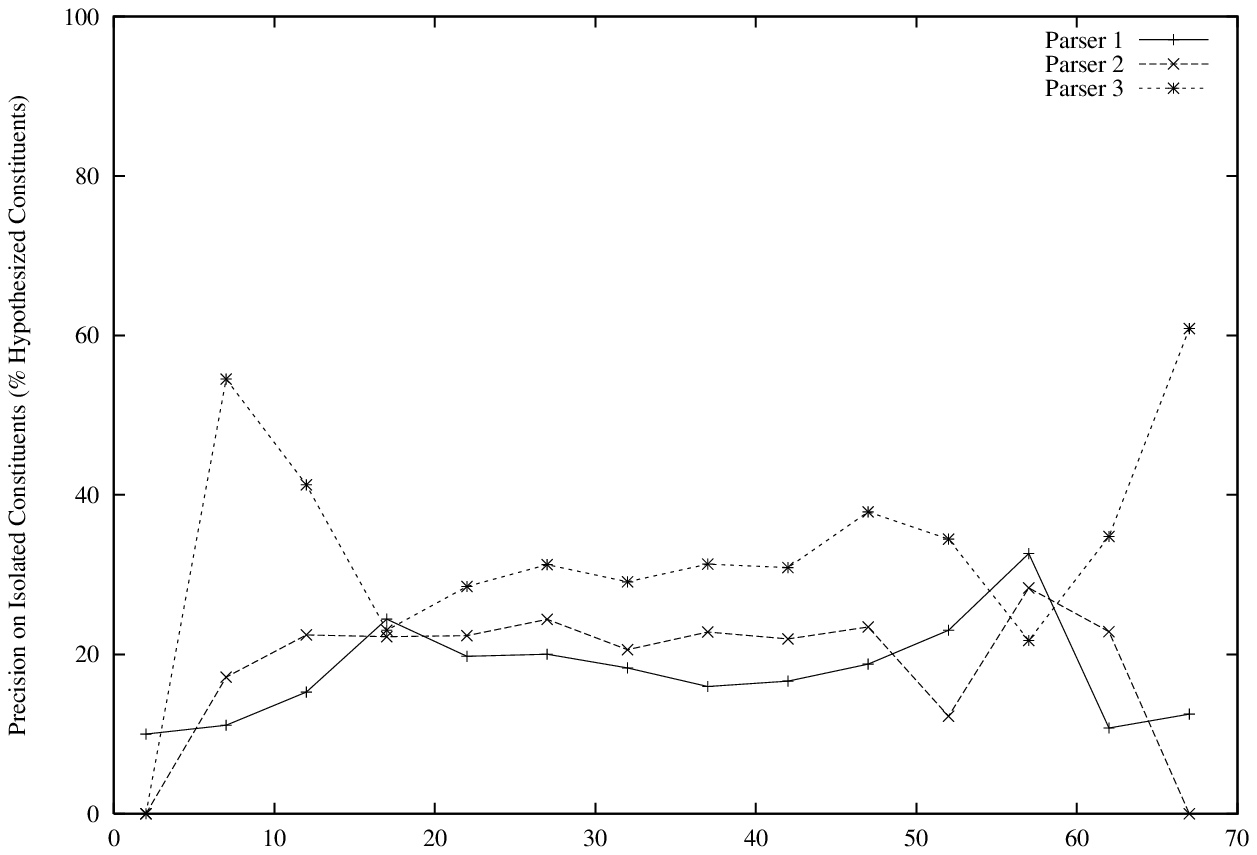, height=0.4\textheight}
    \epsfig{file=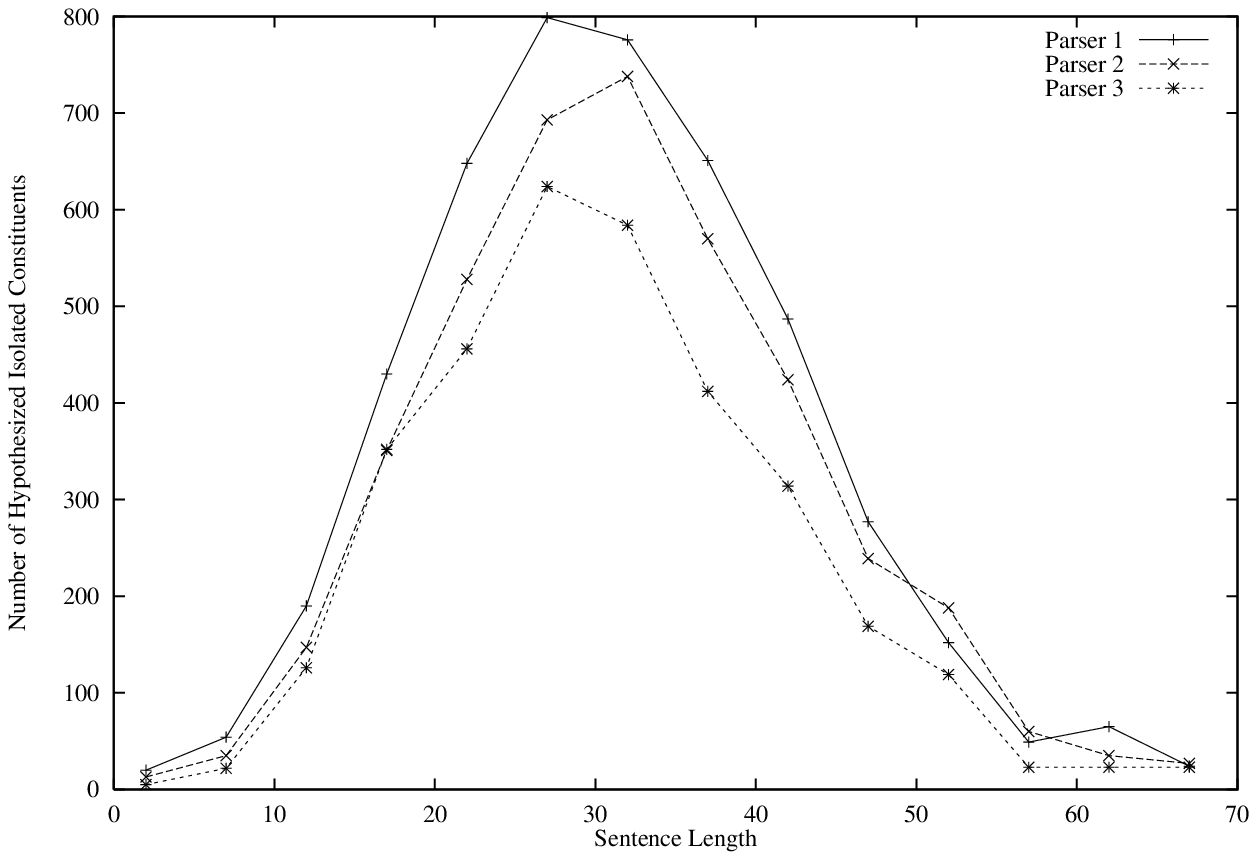, height=0.4\textheight}
  \end{center}
  \caption{Isolated Constituent Precision and Sentence Length}
  \label{fig:isolated:slength}
\end{figure*}
\begin{figure*}
  \begin{center}
    \epsfig{file=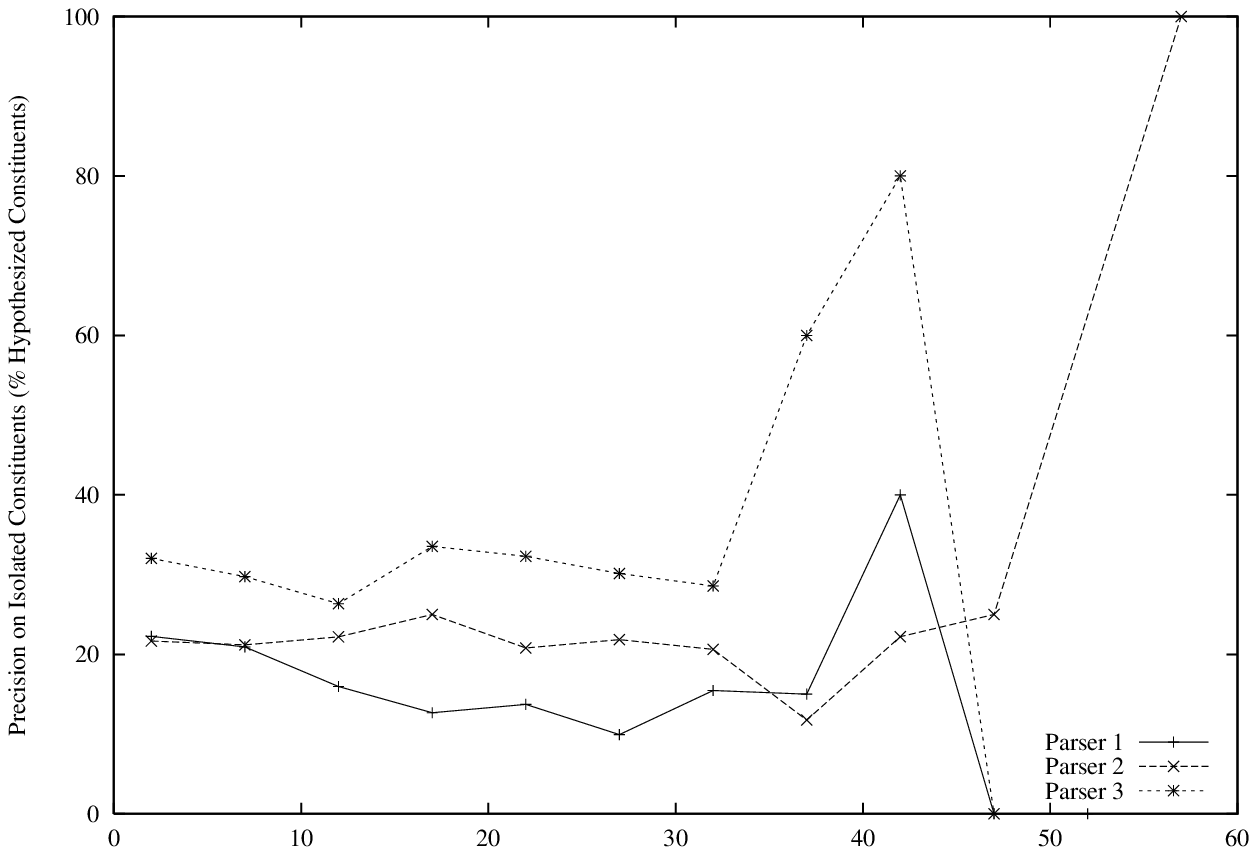, height=0.4\textheight}
    \epsfig{file=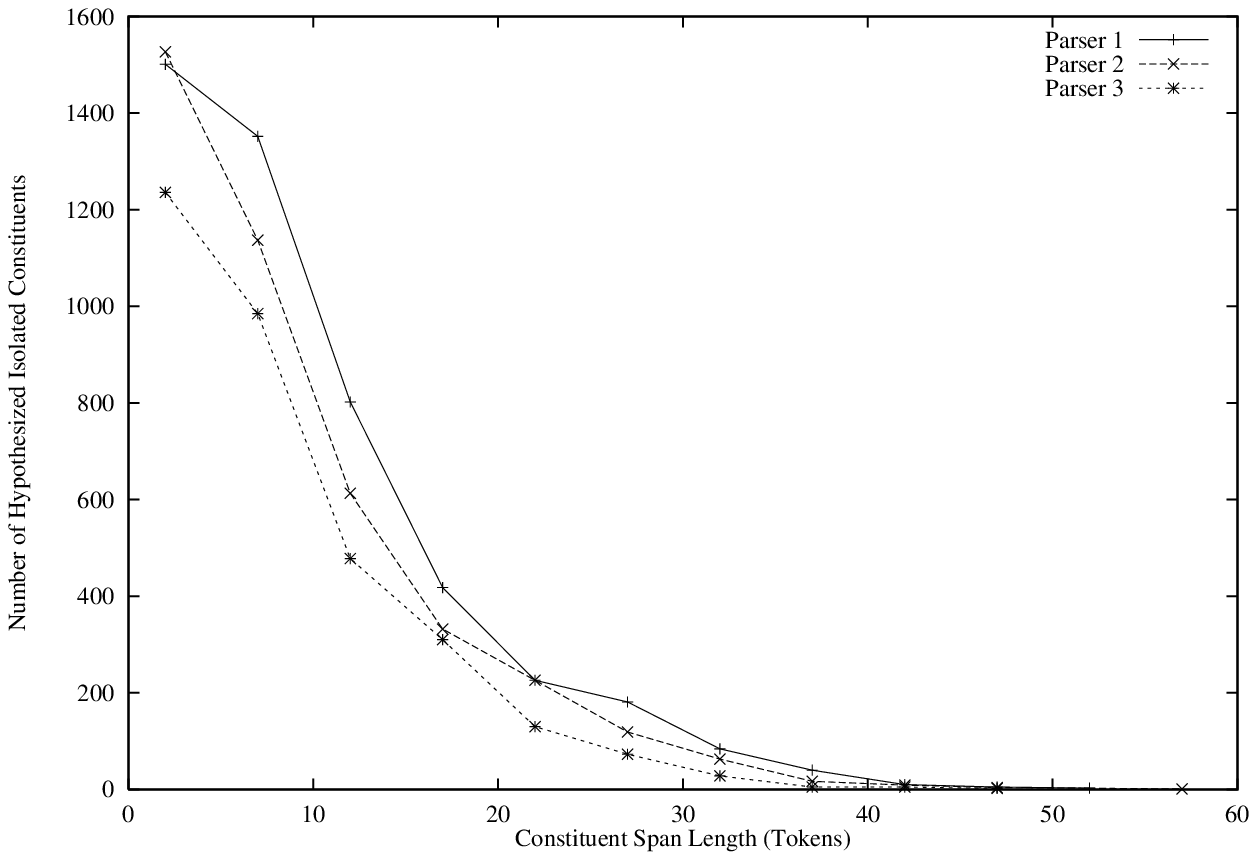, height=0.4\textheight}
  \end{center}
  \caption{Isolated Constituent Precision and Span Length}
  \label{fig:isolated:clength}
\end{figure*}

We show the results of three of the experiments we conducted to
measure isolated constituent precision under various partitioning
schemes.  In Table \ref{table:isolated:tag} we see with very few
exceptions that the isolated constituent precision is less than 0.5
when we use the constituent label as a feature.  The counts represent
portions of the approximately 44000 constituents hypothesized by the
parsers in the development set.  In the cases where isolated
constituent precision is larger than 0.5 the affected portion of the
hypotheses is negligible.

Similarly Figures \ref{fig:isolated:slength} and
\ref{fig:isolated:clength} show how the isolated constituent precision
varies by sentence length and the size of the span of the hypothesized
constituent.  In each figure the upper graph shows the isolated
constituent precision and the bottom graph shows the corresponding
number of hypothesized constituents.  Again we notice that the
isolated constituent precision is larger than 0.5 only in those
partitions that contain very few samples.  From this we see that a
finer-grained model for parser combination, at least for the features
we have examined, will not give us any additional power.

\subsection{Performance Testing}

\begin{table*}
  \begin{center}
    \begin{tabular}{|l|rr|rr|}
      \hline 
      Reference / System 
      &\multicolumn{1}{c}{P} 
      & \multicolumn{1}{c|}{R} 
      & \multicolumn{1}{c}{(P+R)/2} 
      & \multicolumn{1}{c|}{F} \\
      \hline 
      Average Individual Parser & 87.14 & 86.91 & 87.02 & 87.02 \\
      Best Individual Parser    & 88.73 & 88.54 & 88.63 & 88.63 \\
      \hline
      Parser Switching Oracle   & 93.12 & 92.84 & 92.98 & 92.98 \\
      Maximum Precision Oracle  & 100.00& 95.41 & 97.70 & 97.65 \\
      \hline
      Similarity Switching      & 89.50 & 89.88 & 89.69 & 89.69 \\
      \hline
      Constituent Voting        & 92.09 & 89.18 & 90.64 & 90.61 \\
      \hline 
    \end{tabular}
    \caption{Summary of Development Set Performance}
    \label{table:trainingset}
  \end{center}
\end{table*}

The results in Table \ref{table:trainingset} were achieved on the
development set.  The first two rows of the table are baselines.  The
first row represents the average accuracy of the three parsers we
combine. The second row is the accuracy of the best of the three
parsers.\footnote{ The identity of this parser is not given, nor is
  the identity disclosed for the results of any of the individual
  parsers.  We do not aim to compare the performance of the individual
  parsers, nor do we want to bias further research by giving the
  individual parser results for the test set.  }  The next two rows
are results of oracle experiments.  The parser switching oracle is the
upper bound on the accuracy that can be achieved on this set in the
parser switching framework.  It is the performance we could achieve if
an omniscient observer told us which parser to pick for each of the
sentences.  The maximum precision row is the upper bound on accuracy
if we could pick exactly the correct constituents from among the
constituents suggested by the three parsers. Another way to interpret
this is that less than 5\% of the correct constituents are missing
from the hypotheses generated by the union of the three parsers.  The
maximum precision oracle is an upper bound on the possible gain we can
achieve by parse hybridization.

We do not show the numbers for the Bayes models in Table
\ref{table:trainingset} because the parameters involved were
established using this set.  The precision and recall of similarity
switching and constituent voting are both significantly better than
the best individual parser, and constituent voting is significantly
better than parser switching in precision.\footnote{All significance
  claims are made based on a binomial hypothesis test of equality
  with an $\alpha < 0.01$ confidence level.} Constituent voting gives
the highest accuracy for parsing the Penn Treebank reported to date.


\begin{table*}
  \begin{center}
    \begin{tabular}{|l|rr|rr|}
      \hline 
      Reference / System 
      &\multicolumn{1}{c}{P} 
      & \multicolumn{1}{c|}{R} 
      & \multicolumn{1}{c}{(P+R)/2} 
      & \multicolumn{1}{c|}{F} \\
      \hline 
      Average Individual Parser & 87.61 & 87.83 & 87.72 & 87.72\\
      Best Individual Parser    & 89.61 & 89.73 & 89.67 & 89.67\\
      \hline
      Parser Switching Oracle   & 93.78 & 93.87 & 93.82 & 93.82\\
      Maximum Precision Oracle  & 100.00& 95.91 & 97.95 & 97.91\\
      \hline
      Similarity Switching   & 90.04 & 90.81 & 90.43 & 90.43\\
      Bayes Switching           & 90.78 & 90.70 & 90.74 & 90.74\\
      \hline 
      Constituent Voting        & 92.42 & 90.10 & 91.26 & 91.25\\
      Na\"ive Bayes               & 92.42 & 90.10 & 91.26 & 91.25\\
      \hline
    \end{tabular}
    \caption{Test Set Results}
    \label{table:testset}
  \end{center}
\end{table*}

Table \ref{table:testset} contains the results for evaluating our
systems on the test set (section 22).  All of these systems were run
on data that was not seen during their development.  The difference in
precision between similarity and Bayes switching techniques is
significant, but the difference in recall is not.  This is the first
set that gives us a fair evaluation of the Bayes models, and the Bayes
switching model performs significantly better than its non-parametric
counterpart.  The constituent voting and na\"ive Bayes techniques are
equivalent because the parameters learned in the training set did not
sufficiently discriminate between the three parsers.

\begin{table}
  \begin{center}
    \begin{tabular}{|l|r|r|}
      \hline
      \multicolumn{1}{|c|}{Parser}
      &
      \multicolumn{1}{c|}{Sentences}
      &
      \multicolumn{1}{c|}{\%}
      \\
      \hline
      Parser 1 & 279  & 16\\
      Parser 2 & 216  & 13 \\
      Parser 3 & 1204 & 71 \\
      \hline
    \end{tabular}
    \caption{Bayes Switching Parser Usage}
    \label{table:switchingusage}
  \end{center}
\end{table}

Table \ref{table:switchingusage} shows how much the Bayes switching
technique uses each of the parsers on the test set.  Parser 3, the
most accurate parser, was chosen 71\% of the time, and Parser 1, the
least accurate parser was chosen 16\% of the time.  Ties are rare in
Bayes switching because the models are fine-grained -- many
estimated probabilities are involved in each decision.

\subsection{Robustness Testing}

\begin{table*}
  \begin{center}
    \begin{tabular}{|l|rr|rr|}
      \hline 
      Reference / System 
      &\multicolumn{1}{c}{P} 
      & \multicolumn{1}{c|}{R} 
      & \multicolumn{1}{c}{(P+R)/2} 
      & \multicolumn{1}{c|}{F} \\
      \hline 
      Average Individual Parser & 84.55 & 80.91 & 82.73 & 82.69\\
      Best Individual Parser    & 89.61 & 89.73 & 89.67 & 89.67\\
      \hline
      Parser Switching Oracle   & 93.92 & 93.88 & 93.90 & 93.90\\
      Maximum Precision Oracle  & 100.00& 96.66 & 98.33 & 98.30\\
      \hline
      Similarity Switching      & 89.90 & 90.89 & 90.40 & 90.39\\
      Bayes Switching           & 90.94 & 90.70 & 90.82 & 90.82\\
      \hline 
      Constituent Voting        & 89.78 & 91.80 & 90.79 & 90.78\\
      Na\"ive Bayes             & 92.42 & 90.10 & 91.26 & 91.25\\
      \hline
    \end{tabular}
    \caption{Robustness Test Results}
    \label{table:robusttestset}
  \end{center}
\end{table*}

In the interest of testing the robustness of these combining
techniques, we added a fourth, simple non-lexicalized PCFG parser.
The PCFG was trained from the same sections of the Penn Treebank as
the other three parsers.  It was then tested on section 22 of the
Treebank in conjunction with the other parsers.

The results of this experiment can be seen in Table
\ref{table:robusttestset}.  The entries in this table can be compared
with those of Table \ref{table:testset} to see how the performance of
the combining techniques degrades in the presence of an inferior
parser.  As seen by the drop in average individual parser performance
baseline, the introduced parser does not perform very well.  The
average individual parser accuracy was reduced by more than 5\% when
we added this new parser, but the precision of the constituent voting
technique was the only result that decreased significantly.  The Bayes
models were able to achieve significantly higher precision than their
non-parametric counterparts.  We see from these results that the
behavior of the parametric techniques are robust in the presence of
a poor parser.  Surprisingly, the non-parametric switching
technique also exhibited robust behaviour in this situation.

\section{Conclusion}

We have presented two general approaches to studying parser
combination: parser switching and parse hybridization.  For each
experiment we gave an non-parametric and a parametric technique for
combining parsers.  All four of the techniques studied result in
parsing systems that perform better than any previously reported.
Both of the switching techniques, as well as the parametric
hybridization technique were also shown to be robust when a poor
parser was introduced into the experiments.  Through parser
combination we have reduced the precision error rate by 30\% and the
recall error rate by 6\% compared to the best previously published
result.

Combining multiple highly-accurate independent parsers yields
promising results.  We plan to explore more powerful techniques for
exploiting the diversity of parsing methods.

\section{Acknowledgements}
We would like to thank Eugene Charniak, Michael Collins, and Adwait
Ratnaparkhi for enabling all of this research by providing us with
their parsers and helpful comments.

This work was funded by NSF grant IRI-9502312.  Both authors are
members of the Center for Language and Speech Processing at Johns
Hopkins University.

\bibliographystyle{acl}
\bibliography{emnlp99henderson}

\end{document}